\documentclass[10pt,twocolumn,letterpaper]{article}

\usepackage[pagenumbers]{cvpr} 

\usepackage{graphicx}
\usepackage{CJKutf8}
\usepackage{amsmath}
\usepackage{amssymb}
\usepackage{booktabs}

\usepackage[pagebackref,breaklinks,colorlinks]{hyperref}

\usepackage[capitalize]{cleveref}
\crefname{section}{Sec.}{Secs.}
\Crefname{section}{Section}{Sections}
\Crefname{table}{Table}{Tables}
\crefname{table}{Tab.}{Tabs.}

\def\cvprPaperID{*****} 
\def\confName{CVPR}
\def\confYear{2024}

\begin{document}
\begin{CJK}{UTF8}{gbsn}

\title{Semi-supervised Video Semantic Segmentation Using Unreliable Pseudo Labels for PVUW2024}
\author{
Biao Wu\and
Diankai Zhang\and
Si Gao\and
Chengjian Zheng\and
Shaoli Liu\and
Ning Wang\and
State Key Laboratory of Mobile Network and Mobile Multimedia Technology,ZTE,China
\and
{\tt\small \{wu.biao,zhang.diankai,gao.si,zheng.chengjian,liu.shaoli,wangning\}@zte.com.cn}
}
\maketitle

\begin{abstract}
   Pixel-level Scene Understanding is one of the fundamental problems in computer vision, which aims at recognizing object classes, masks and semantics of each pixel in the given image. Compared with image scene parsing, video scene parsing introduces temporal information, which can effectively improve  the  consistency  and  accuracy  of  prediction,because the real-world is actually video-based rather than a static state. In this paper, we adopt semi-supervised video semantic segmentation method based on unreliable pseudo labels. Then, We ensemble the teacher network model with the student network model to generate pseudo labels and retrain the student network. Our method achieves the mIoU scores of 63.71\% and 67.83\% on development test and final test respectively. Finally, we obtain the 1st place in the Video Scene Parsing in the Wild Challenge at CVPR 2024.
\end{abstract}


\section{Introduction}
\label{sec:intro}

The Video Scene Parsing in the Wild (VSPW) \cite{miao2021vspw} is a video semantic segmentation dataset with 3536 videos and annotations of 124 categories .Thanks to the availability of various semantic segmentation datasets, significant progress has been made in image semantic segmentation. The challenge aims to assign pixel-wise semantic labels to each video frame of the test set videos in the VSPW. The prominent evaluation metric for the challenge is the mean Intersection over Union (mIoU).

\begin{figure*}[ht]
    \centering
    \includegraphics[width= 16cm]{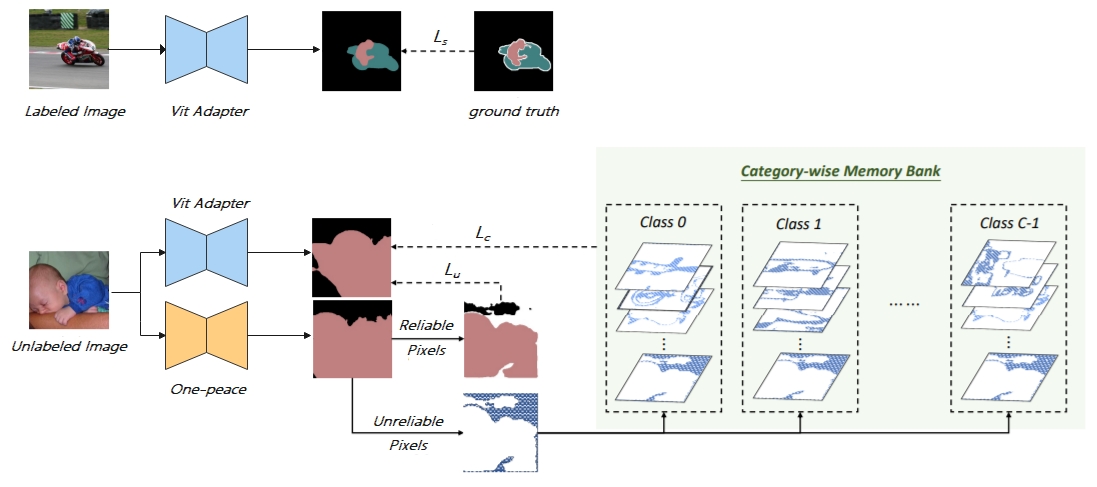}
    \caption{The overall architecture of our method.}
    \label{fig:my_diagram}
\end{figure*}

With the development of deep neural networks and the availability of large-scale labeled data, the capabilities of Video Semantic Segmentation (VSS) have been notably extended \cite{Agronav}. VSS is a spatial-temporal variation of image segmentation on videos that aims to predict a pixel-wise label across sequential video frames\cite{Zero-Shot}.Compared with image semantic segmentation, most existing methods of VSS methods emphasize the exploitation of temporal information.Several approaches \cite{Lup,Zhu,Gadde} utilized optical flow prediction to model temporal information between frames.However, optical flow may lead to unbalanced latency. ETC\cite{Liu} used the temporal loss and the new temporal consistency knowledge distillation on the per-frame segmentation predictions as an efficient replacement for optical flow. MRCFA \cite{Sun} mining the relationship of cross-frame affinities to achieve better temporal information aggregation.TMANet \cite{Wang} is the first work using a temporal memory attention module to capture the temporal relations between frames in VSS. LLVSS \cite{Li} designed an efficient framework involving both adaptive feature propagation and adaptive key-frame schedule.DVIS \cite{DVIS} streamlines the process by framing VSS as an initial segmentation task followed by tracking, subsequently refining segmentation outcomes using comprehensive video data. The 1st Place Solution for CVPR 2023 PVUW VSS Track \cite{Yan} focusing on enhancing spatial-temporal correlations with contrastive loss and leveraging multi-dataset training with label mapping to boost model performance. 
\section{Approach}
\label{sec:Approach}
In this section, we describe the overall architecture of our network. And then, we introduce a semi-supervised
video semantic segmentation method based on unreliable pseudo labels\cite{wang2022semi}. We first train the teacher network and student network on labeled data, then use the teacher network to generate pseudo labels, combine them with the original dataset to form a new dataset, and then retrain the student network. Through semi-supervised training, we improve the performance of the model on unlabeled datasets.

\subsection{Overview}
Transformer is a neural network model based on attention mechanism, which has achieved significant success in natural language processing and other sequence data processing tasks. In recent years, with the development of transformer technology, it has also made significant progress in the field of segmentation. Given that the One-peace \cite{wang2023one}algorithm achieves state-of-the-art performance in the field of semantic segmentation  on ADE20k\cite{zhou2017scene} dataset, we choose it as the teacher network. Meanwhile, we choose the ViT-Adapter\cite{chen2022vitadapter} algorithm as the student network.

\subsection{Semi-supervised method }
With the development of deep learning methods, there is a qualitative improvement in segmentation performance. However, high-performance deep learning models require a large amount of data and labeling, especially pixel level labeling, which requires a huge investment of manpower and time costs. Therefore, methods based on semi-supervised learning are highly favored by researchers. The core problem of semi-supervised learning is to effectively utilize unlabeled samples as a supplement to labeled samples, thereby improving the performance of the model. The conventional semi- supervised methods retain high confidence prediction results by screening samples, but these results in a large amount of unlabeled data not being effectively utilized, leading to insufficient model training. It is difficult to assign the correct labels to unlabeled pixels for some unpredictable categories.
Therefore, we treat unreliable prediction results as negative samples to participate in model training, allowing all unlabeled samples to play an effective role in the training process.


\subsection{Pseudo-labeling strategy}
To avoid overfitting incorrect pseudo labels, we use the probability distribution entropy of each pixel to filter high-quality pseudo labels.  Specifically, we denote $p_{ij}\in R^C$ as the softmax probabilities generated by the segmentation head for the \emph{i}-th unlabeled image at pixel \emph{j}, where \emph{C} is the number of
classes.

\begin{equation}
\begin{aligned}
H(p_{ij})=-\sum_{c=0}^{C-1}p_{ij}(c)\log p_{ij}(c)
\end{aligned}
\end{equation}

where $p_{ij}(c)$  is the value of $p_{ij}$ at \emph{c}-th dimension.
We define the pseudo-label for the \emph{i}-th unlabeled image at
pixel \emph{j} as:

\begin{equation}
\begin{split}
\hat{y}{_{ij}^{u}}= \left \{
\begin{array}{ll}
   \underset{c}{{\arg\max}\, p_{ij}(c) },                    & if H(p_{ij}) < \gamma_t\\
    ignore,                                 & otherwise
\end{array}
\right.
\end{split}
\end{equation}

We use pixel level entropy to distinguish between reliable and unreliable pixels in pseudo labels.

\subsection{The pipeline of our method}
As shown in Figure 1, how to
extract effective information from unmarked data is a crucial factor, so we use semi-supervised learning methods. Specificly,in the first step, we train the teacher network model and student network model using labeled training data, and then use multi-scale and horizontal flipping to enhance testing and model ensemble to generate pseudo labels. Then we combine the unlabeled and labeled datasets into a new dataset, and continue fine-tuning the student network model. For pseudo labels, pixel level entropy is used to filter reliable pixels and unreliable pixels. For unreliable pixels as negative samples, comparative loss training is used to ensure that the entire pseudo label can be effectively used during the training process.

\subsection{Loss}
For every labeled image, our goal is to minimize
the standard cross-entropy loss in Eq(4). For each unlabeled image, we first use a teacher model for prediction. Then, we use pixel level entropy to ignore unreliable pixel level pseudo labels and unsupervised losses in Eq(5). We use contrastive loss\cite{contrastive} to fully utilize the excluded unreliable pixels in Eq(6).
To achieve better segmentation performance, we minimize overall loss to the greatest extent possible,
it can be formalized as:
\begin{equation}
\begin{aligned}
L=L_s + \lambda_u L_u + \lambda_cL_c
\end{aligned}
\end{equation}

where $L_s$ and $L_u$ represent the standard cross-entropy loss, $L_c$ is the contrastive loss.

\begin{equation}
\begin{aligned}
L_s=\frac{1}{|B_l|}\sum_{(x_i^l,y_i^l)\in B_l}l_{ce}(f\circ h(x_i^l;\theta), y_i^l)
\end{aligned}
\end{equation}

\begin{equation}
\begin{aligned}
L_u=\frac{1}{|B_u|}\sum_{x_i^u\in B_u}l_{ce}(f\circ h(x_i^u;\theta), \hat{y}_i^u)
\end{aligned}
\end{equation}

\begin{multline}
L_c=-\frac{1}{C\times M}\sum_{c=0}^{C-1} \\ 
\log [\frac{e^{\left\langle z_{ci}, z_{ci}^+ \right\rangle}/\tau}{e^{\left\langle z_{ci}, z_{ci}^+ \right\rangle}/\tau + \sum_{j=1}^{N}e^{\left\langle z_{ci}, z_{ij}^- \right\rangle}/\tau}]
\end{multline}

\section{Experiments}
\label{sec:Experiments}

In this part, we will describe the implementation details of our proposed method and report the results on the PVUW2024  challenge test set.

\begin{table}
  \centering
  \resizebox{0.67\columnwidth}{!}{
  \begin{tabular}{c|cccc}
    \toprule
    Method & Backbone &mIoU \\
    \hline
Mask2Former&Swin-L	&0.5709	\\
Mask2Former&BEiT-L	&0.5854	\\
Mask2Former&ViT-Adapter-L&0.6140\\
Mask2Former&One-peace&0.6196\\
    \bottomrule
  \end{tabular}
  }
  \caption{Experiments of different backbones on PVUW2024 challenge test part 1.}
  \label{tab:1}
\end{table}

\begin{table}
  \centering
  \resizebox{0.6\columnwidth}{!}{
  \begin{tabular}{c|cccc}
    \toprule
    Method&Crop Size & mIoU \\
    \hline
ViT-Adapter-L&224	&0.5795	 \\
ViT-Adapter-L&320	&0.5890	 \\
ViT-Adapter-L&640	&0.6140	  \\
ViT-Adapter-L&768	&0.6169	 \\
ViT-Adapter-L&896	&0.6203	  \\
    \bottomrule
  \end{tabular}
  }
  \caption{Experiments of different crop size with ViT-Adapter-L on PVUW2024 challenge test part 1.}
  \label{tab:2}
\end{table}

\begin{table*}
  \centering
  \begin{tabular}{cc|ccc|ccc}
    \toprule
    Method & crop size & multiscale & flip & Semi-supervised training &	mIoU	&VC8	&VC16 \\
    \hline
    ViT-Adapter-L    &896&\checkmark &\checkmark & 	&0.6236	&0.9330	&0.9097 \\
    $\star$ ViT-Adapter-L &896&\checkmark &\checkmark &\checkmark &0.6276	&0.9432	&0.9214 \\
    One-peace    &896&\checkmark &\checkmark & 	&0.6294	&0.9433	&0.9219 \\
    $\star$ One-peace &896&\checkmark &\checkmark &\checkmark &0.6338	&0.9434	&0.9218 \\
    Ensemble($\star$)&   &           &           &           &0.6371	&0.9475	&0.9269 \\
    \bottomrule
  \end{tabular}
  \caption{Experiments of Inference Augmentation on PVUW2024 challenge test part 1.}
  \label{tab:example}
\end{table*}

\subsection{Datasets}

The VSPW Dataset annotates 124 categories of real-world scenarios, which contains 3,536 videos, with 251,633 frames totally. Among these videos, there are 2806 videos in the training set, 343 videos in the validation set, and 387 videos in the testing set. In order to enrich our training samples, both the training and validation set are used for training. Due to the large number of parameters in the transformer model, increasing the number of training samples is beneficial for improving the performance of the model. We introduce additional data to train our model, such as the ADE20k\cite{zhou2017scene} and COCO\cite{lin2014microsoft} datasets. During the training phase, the backbone of our model is pretrained on the ImageNet22K\cite{deng2009imagenet} dataset. The COCO dataset is used to train the entire model during the pre-training phase. The COCO and ADE20k dataset labels are mapped to the VSPW dataset through label remapping, and categories that do not exist in the VSPW dataset are marked as 255.

\subsection{Training Configuration}
All models in our experiments are implemented in the PyTorch\cite{paszke2017automatic} framework. For data augmentation, we perform random resizing within ratio range [0.5, 2.0], random cropping, random horizontal flipping, and color jitter on images. An AdamW optimizer is applied with the initial learning rate of 1e-5, beta = (0.9, 0.999), and the weight decay of 0.05. A linear warmup is used. To prevent overfitting, the training iterations is set to 80k.

\begin{table}
  \centering
  \resizebox{1.0\columnwidth}{!}{
  \begin{tabular}{ccccc}
    \toprule
    Team & mIoU & Weight IoU & VC8 & VC16 \\
    \midrule
    SiegeLion&	0.6783& 0.7761& 0.9482& 0.9290  \\
    lieflat& 0.6727& 0.7659& 0.9499& 0.9312 \\
    kevin1234& 0.6392& 0.7447 & 0.9484& 0.9325 \\
    bai\_kai\_shui& 0.6375& 0.7422& 0.9462& 0.9290 \\
    JMCarrot& 0.6337& 0.7456& 0.9462& 0.9294 \\
    ipadvideo& 0.5854 & 0.7145& 0.9073& 0.8802 \\
    \bottomrule
  \end{tabular}
  }
  \caption{Comparison with other teams on the PVUW2024 challenge final test set.}
  \label{tab:example}
\end{table}

\begin{figure*}[ht]
    \centering
    \includegraphics[width= 16cm]{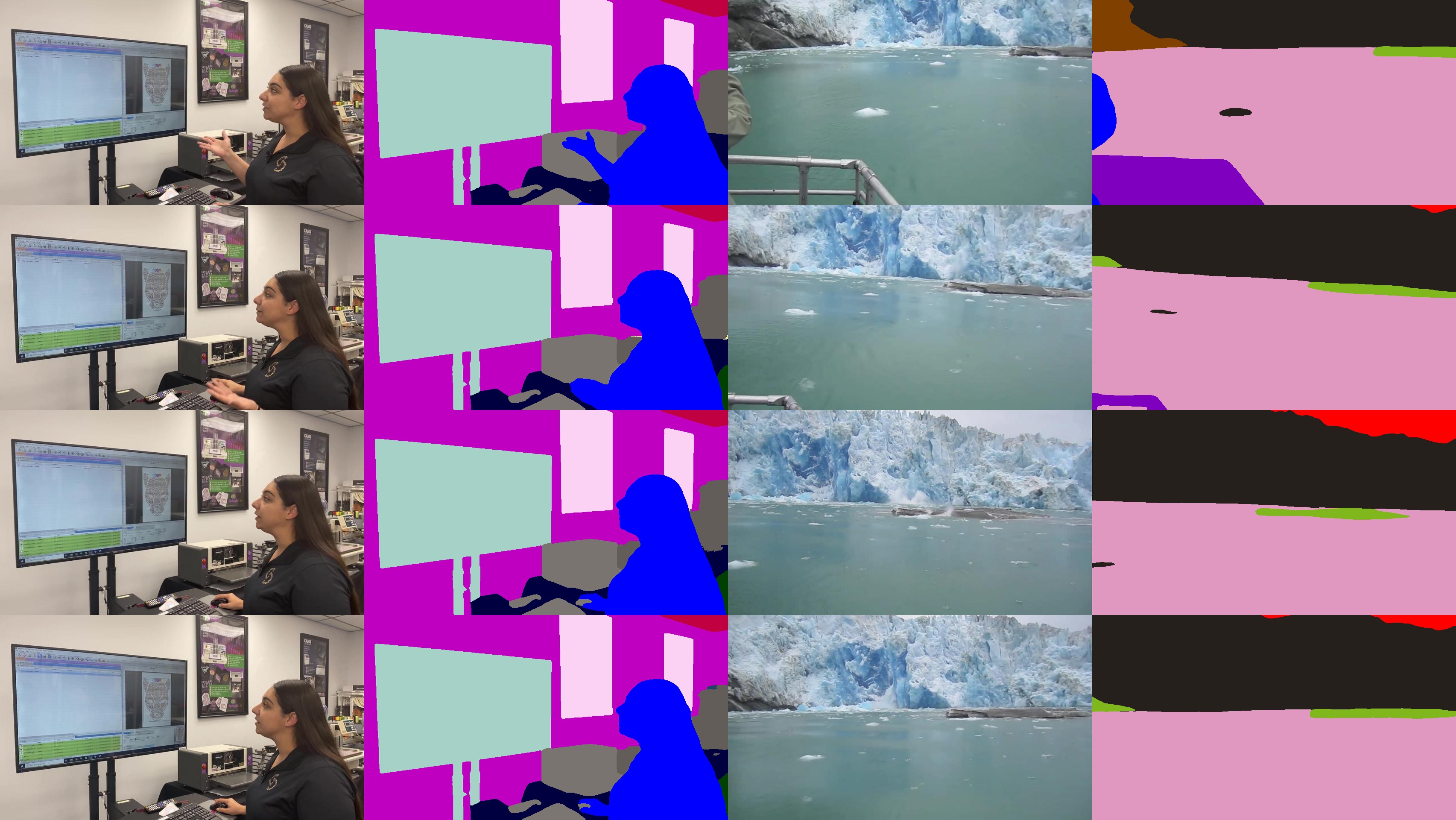}
    \caption{Qualitative result on VSPW\cite{miao2021vspw} test set of out method.}
    \label{fig:my_diagram}
\end{figure*}

\subsection{Ablation Studies}
With the rapid development of transformer technology, models based on transformer technology have shown strong feature expression ability in the field of dense object detection and segmentation, and even maintain good robustness in some complex scenes. Therefore, we explore the application of models based on transformer in video semantic segmentation tasks. The experimental results of different backbones and methods are shown in Table 1. From the table, it can be seen that selecting One-peace as the backbone has significantly better performance than Swin-L\cite{liu2021swin}, BEiT-L\cite{bao2021beit} and ViT-Adapter-L\cite{chen2022vitadapter}. 
In subsequent experiments, we continue to explore the effects of network input resolution, multi-scale and flip enhancement testing, semi-supervised training, and model ensemble on segmentation performance.

\subsection{Semi-supervised training}
Semi-supervised learning aims to extract effective information from unlabeled data, thereby improving the performance of the model.
Taking inspiration from this, we chose one-place as the teacher network and ViT-Adapter as the student network. Firstly, we train teacher and student networks on labeled datasets, and generate pseudo labels through multi-scale and flip enhanced testing and model fusion. We combine unlabeled and labeled datasets into a new training set to continue fine-tuning the student network. We believe that in semi-supervised model training, each pixel of the pseudo label is important, even if its prediction is ambiguous. Intuitively, unreliable predictions may be directly confused in the category with the highest probability, but they should have credibility for pixels that do not belong to other categories. Therefore, such pixels can be judged as negative samples in the least likely category. From Table 3, Semi-supervised training and model ensemble improve the mIoU by approximately 0.4 percentage
points.

\subsection{Inference Augmentation}

In the inference stage, we continue to explore the factors that affect model performance. We achieve higher scores on the mIOU metric by using multi-scale and horizontal flipping for each scale where the selected scales are [512./896., 640./896., 768./896., 896./896., 1024./896., 1152./896., 1280./896., 1408./896.]. From Tables 2 and 3, it can be seen that The multi-scale and horizontal flipping results increase the mIoU indicator by 0.4 percentage points compared to the single scale results. In order to further improve the performance of the model, we ensemble the teacher model and student model with a crop size of 896 and achieve the highest score on the mIOU of PVUW test part 1.  We achieve state-of-the-art results on the final test set of the PVUW semantic segmentation challenge by combining multi-scale and horizontal flipping enhancement testing, semi-supervised training learning, and multi model ensemble techniques. Finally, we obtain the 1st place in the final test set, as shown in Table 4. Qualitative result on VSPW test set of our method is shown in Figure 2.

\section{Conclusion}
\label{sec:Conclusion}

In this paper, we start by selecting a strong baseline model that is well-suited for the task of multi-class semantic segmentation. We adopt a semi-supervised video semantic segmentation method based on unreliable pseudo labels. Our method effectively utilizes unlabeled samples as a supplement to labeled samples to improve model performance. We propose a ensemble method to get more accurate probability by fusing the results of different models. These techniques are combined to create a comprehensive solution that achieve 1st place in the VSS track of PVUW challenge at CVPR 2024 conference. The results demonstrate the effectiveness and versatility of our solution for addressing multi-task semantic segmentation problems.

\end{CJK}
\end{document}